\documentclass{article}

\usepackage{microtype}
\usepackage{graphicx}
\usepackage{subfigure}
\usepackage{booktabs} 
\usepackage{apacite}
\usepackage{bbm}

\usepackage{hyperref}

\usepackage[accepted]{icml2025}

\usepackage{amsmath}
\usepackage{amssymb}
\usepackage{mathtools}
\usepackage{amsthm}

\usepackage[capitalize,noabbrev]{cleveref}

\theoremstyle{plain}

\theoremstyle{definition}

\theoremstyle{remark}

\usepackage[textsize=tiny]{todonotes}

\icmltitlerunning{Active Concept Bottleneck Models for Interpretable RLHF}

\usepackage{amsmath,amsfonts,bm}

\def\eqref#1{equation~\ref{#1}}

\def\1{\bm{1}}

\DeclareMathAlphabet{\mathsfit}{\encodingdefault}{\sfdefault}{m}{sl}
\SetMathAlphabet{\mathsfit}{bold}{\encodingdefault}{\sfdefault}{bx}{n}

\def\gX{{\mathcal{X}}}
\def\gY{{\mathcal{Y}}}

\begin{document}

\twocolumn[
\icmltitle{Interpretable Reward Modeling with Active Concept Bottlenecks }



\icmlsetsymbol{equal}{*}

\begin{icmlauthorlist}
\icmlauthor{Sonia Laguna}{yyy}
\icmlauthor{Katarzyna Kobalczyk}{comp}
\icmlauthor{Julia E. Vogt}{yyy}
\icmlauthor{Mihaela van der Schaar}{comp}
\end{icmlauthorlist}

\icmlaffiliation{yyy}{Department of
Computer Science, ETH Zurich, Zurich, Switzerland}
\icmlaffiliation{comp}{Department of Applied Mathematics and Theoretical Physics,
University of Cambridge, United Kingdom}

\icmlcorrespondingauthor{Sonia Laguna}{slaguna@inf.ethz.ch}
\icmlcorrespondingauthor{Katarzyna Kobalczyk}{knk25@cam.ac.uk}

\icmlkeywords{Machine Learning, ICML}

\vskip 0.3in
]



\printAffiliationsAndNotice{\icmlEqualContribution} 

\begin{abstract}
We introduce Concept Bottleneck Reward Models (CB-RM), a reward modeling framework that enables interpretable preference learning through selective concept annotation. Unlike standard RLHF methods that rely on opaque reward functions, CB-RM decomposes reward prediction into human-interpretable concepts. To make this framework efficient in low-supervision settings, we formalize an active learning strategy that dynamically acquires the most informative concept labels. We propose an acquisition function based on Expected Information Gain and show that it significantly accelerates concept learning without compromising preference accuracy. Evaluated on the UltraFeedback dataset, our method outperforms baselines in interpretability and sample efficiency, marking a step towards more transparent, auditable, and human-aligned reward models.
\end{abstract}

\section{Introduction}

\paragraph{Motivation.} A key challenge in aligning machine learning (ML) systems, particularly Large Language Models (LLM), with human preferences lies in the interpretability of the reward models used for their alignment with human values. While significant progress has been made in learning reward functions from human feedback~\citep{christiano2017deep, ouyang2022training}, most existing approaches rely on black-box reward models, making it difficult to understand which factors drive human preferences. This lack of interpretability limits the ability to diagnose, refine, and trust these models in real-world applications ~\citep{doshi2017towards}. To address this, we propose a framework for learning interpretable reward functions that explicitly identify and leverage latent dimensions of human preferences with minimal annotation costs, building on the Concept Bottleneck Model (CBM) paradigm~\citep{koh2020concept}. By actively querying human feedback for annotations of interpretable latent dimensions, we aim to uncover the underlying concepts that influence user decisions. Our proposed approach not only enhances the transparency of reward models but also ensures robust alignment with human values. Integrating uncertainty estimation and bayesian experimental design~\citep{melodeep}, we optimize the human feedback, scaling the annotation process. This work advances the development of interpretable and trustworthy systems, with more transparent human-machine interactions.

\paragraph{Background}

Reward modeling is central to Reinforcement Learning from Human Feedback (RLHF), a framework where ML systems learn to align with human preferences by optimizing behavior based on feedback—typically in the form of pairwise comparisons between model outputs. Instead of handcrafting reward functions, RLHF uses human preference feedback to train proxy reward models, which are then used to guide policy optimization. This has been effective in fine-tuning LLMs for instruction following and safe interaction tasks~\citep{christiano2017deep, ouyang2022training}. However, the reward models used in RLHF are often opaque and monolithic, making it difficult to understand, debug, or adapt their behavior~\citep{bai2022training}. Moreover, they require large quantities of labeled data, which is expensive and generally noisy~\citep{casper2023open,sharma2024towards}. To faithfully capture the spectrum of human intent, reward models should support interpretability, personalization, and uncertainty estimation. These properties are not only useful for model introspection but are critical for scalable and trustworthy alignment. 

CBMs~\citep{koh2020concept} offer an interpretable alternative to black-box function approximators by explicitly modeling the intermediate concepts that drive model decisions. These structured neural models decompose predictions into two stages: (i) predicting human-interpretable concepts from raw inputs, and (ii) predicting the final task label based solely on these concepts. This structure enables inspection, intervention, and debugging of the model’s behavior. CBMs have been extended to interactive settings with different interaction policies~\citep{chauhan2023interactive}, stochastic variants considering correlations to propagate through predictions~\citep{vandenhirtz2024stochastic}, and adapting to external interventions from pretrained models~\citep{laguna2024beyond}. Although previous work explores how to intervene most effectively at test time, the strategies assume full access to concept annotations during training. This is, however, not the case in typical data collection setups for preference learning. In this work, we introduce an Active Learning (AL) framework on Concept Bottleneck Reward Models (CB-RM). Our AL algorithm must decide which concept labels to query during training for maximal utility. This setting poses new challenges: the model must identify which missing concept labels will have the most influence on generalization and interpretability.  Prioritizing such queries effectively is crucial for building scalable, data-efficient, and trustworthy reward models. A recent attempt at interpretable reward modeling, ArmoRM~\citep{wang2024interpretable}, learns multi-objective concept scores in a similar fashion to CBMs and combines them via a mixture-of-experts gating mechanism. However, we identify several potential concerns regarding this work and address them here.

\paragraph{Contributions} In this work, we overcome the above challenges with the following contributions: (i) We introduce CB-RM, a novel approach for modeling reward functions with underlying representations of human understandable concepts; (ii) We model an AL CB-RM framework that selectively acquires concept annotations during training episodes for improved interpretability and enables effective learning in low-data regimes---we introduce this formalization for the first time in the CBM context, which itself has extensive potential applications; (iii) We demonstrate that active concept acquisition is critical in preference learning settings, where annotating every concept for every pairwise example is expensive and infeasible. To this end, we introduce an acquisition strategy based on Expected Information Gain (EIG), which significantly improves the baselines on concept learning efficiency across training episodes without compromising downstream preference prediction accuracy. Here, we define concepts as interpretable response attributes—such as helpfulness, correctness, and others—that reflect human evaluative preferences.

\begin{figure*}[h!]
\centering
\includegraphics[width=0.95\textwidth]{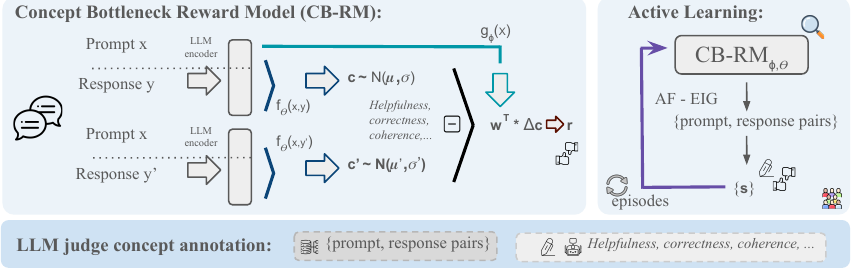}
\vspace*{-0.5em}
\caption{
\textbf{Overview of CB-RM with Active Learning.}
In CB-RM, each prompt-response pair is encoded by an LLM encoder and an MLP to predict Gaussian-distributed concept representations $\Delta c$, to generate a reward prediction using context-conditioned weights $\mathbf{w}$ from $g_\phi(x)$. During Active Learning, concept-label pairs are selectively acquired using an Activation Function (AF), the best performing being EIG, to maximize concept learning across episodes. Human-annotated concept scores from an LLM judge supervise the pipeline.
\vspace*{-0.25cm}
}
\label{fig:overview}
\end{figure*}

\section{Problem Formalism}

\paragraph{Setup} Let $\mathcal{D}_{pool} = \{(x_i, y_i', y_i)\}$ be an unlabeled pool of pairwise preference data conventionally used in reward modeling for RLHF. We denote by $x_i \in \mathcal{X} \subset \Sigma^*$ a prompt, and $y_i', y_i \in \mathcal{Y} \subset \Sigma^*$ two candidate responses. We use $\Sigma^*$ to denote the space of natural language and $\gX$, $\gY$ the subsets of all plausible human prompts and LLM responses, respectively. We assume the existence of a ground-truth human reward function $r: \mathcal{X} \times \mathcal{Y} \rightarrow \mathbb{R}$ that determines the preference choice between two candidate responses based on the standard BTL~\citep{bradley1952rank} model:
\begin{equation}
    p(y \succ y' \vert x) = \sigma\left(r(x, y) - r(x, y')\right).
\end{equation}
In this work, we assume that the rewards assigned to candidate responses depend on a set of underlying latent concepts $\mathcal{C}$ extractable from the texts (i.e. helpfulness, correctness, coherence, ...). Thus, we model $r$ via a \textit{context-aware bottleneck} composed of two functions: $f_\theta$ and $g_\phi$. $f_\theta: \mathcal{X} \times \mathcal{Y} \rightarrow \mathcal{C}$, maps a prompt-response pair $(x, y)$ to a vector $ f_\theta(x, y) \equiv \boldsymbol{c} \in \mathcal{C}$ representing the set of concepts present in $(x, y)$ and $g_\phi: \mathcal{X} \rightarrow \mathbb{R}^K$, where $\dim(\mathcal{C}) = K$, maps the prompt $x$ to weight vector $g_\phi(x) \equiv \boldsymbol{w} \equiv (w_1, w_2, \ldots, w_K) \in \mathbb{R}^K $, with $w_k$ representing the importance of the $k$-th concept in the context of the topic or task implicitly defined in $x$. The final reward function can be described as:
\begin{equation}
    r_{\theta, \phi} (x, y) = g_\phi(x)^Tf_\theta(x, y) = \boldsymbol{w}^T\boldsymbol{c}.
\end{equation}
To account for uncertainty in the concept predictions, we model the concept encoder $f_\theta$ as a probabilistic function. Specifically, for each input $(x, y)$, it predicts a Gaussian distribution over concept scores:
\begin{equation}
    f_\theta(x, y) \sim \mathcal{N}(\boldsymbol{\mu}(x, y), \mathrm{diag}(\boldsymbol{\sigma}^2(x, y))),
\end{equation}
where $\boldsymbol{\mu} \in \mathbb{R}^K$ is the predicted mean vector and $\boldsymbol{\sigma}^2 \in \mathbb{R}^K$ is the vector of variances for each concept dimension. Given two responses $y$ and $y'$ to the same prompt $x$, we define the concept \emph{difference distribution} as the distribution of the vector $\Delta \boldsymbol{c} = f_\theta(x, y) - f_\theta(x, y')$. Since both $f_\theta(x, y)$ and $f_\theta(x, y')$ are modeled as independent Gaussians we have:
{\small
\begin{equation}
    \Delta \boldsymbol{c} \sim \mathcal{N}\left(\boldsymbol{\mu}(x, y) - \boldsymbol{\mu}(x, y'),\ \mathrm{diag}(\boldsymbol{\sigma}^2(x, y) + \boldsymbol{\sigma}^2(x, y'))\right),
\end{equation}
}

which enables downstream acquisition strategies to leverage both mean and variance for uncertainty-aware querying. The CB-RM model is trained jointly on the task and concept objectives, with equal weighting by default. While this framework can be generalized, we focus on classification tasks, using cross-entropy loss for both objectives.

\textbf{Goal} To learn $r_{\theta, \phi}$ we must obtain preference labels $\ell_i \in \{0, 1\}$ determining if for a given sample $(x_i, y_i, y_i')$ the first ($\ell_i = 0$) or the second ($\ell_i = 1$) response is preferred. In addition, to learn the human-interpretable concept bottleneck, we also require a learning signal for grounding the latent representations $\boldsymbol{c}_i$ in a human-interpretable space of concepts. 

In this work, we choose to collect \textit{relative concept labels} $s_i^k \in \{0, 1\}$, where $ i \in \{1, \ldots, |\mathcal{D}_{pool}|\}$ and $k \in \{1, \ldots K\}$. That is, for a tuple $(x_i, y_i, y_i')$, we query binary concept labels indicating which of the two responses $y_i$ or $y_i'$ performs better with respect to the $k$-th concept e.g., which response is more helpful or more coherent. Obtaining both preference labels and concept annotations across the entire dataset $\mathcal{D}_{pool}$—which must be large for robust reward learning—is expensive. Therefore, we propose an \textit{active learning} strategy to efficiently select queries, reducing labeling costs.

\section{Method: Active Learning in CB-RM}

Our method builds on CBMs using probabilistic concept prediction to produce reward scores, and selectively acquires concept labels during training via active learning. In what follows, we formalize this framework and introduce acquisition strategies designed to improve supervision efficiency.

We explore selective acquisition strategies that prioritize concept labels expected to be most beneficial for training. Figure~\ref{fig:overview} shows a general overview of the proposed AL method in CB-RM. To manage training stability and incorporate recent feedback efficiently, we adopt a FIFO (First-In-First-Out)~\cite{dwaracherla2024efficient} replay buffer strategy. After each training episode, newly acquired preference and concept annotations are appended to the buffer, while the oldest entries are discarded once capacity is reached. This approach maintains a bounded memory footprint and ensures the model trains predominantly on up-to-date and relevant samples. Empirically, this strategy supports rapid adaptation to informative queries without overfitting to stale data, aligning with best practices in active exploration under feedback constraints. Algorithm ~\ref{alg:active_cbm}, Appendix~\ref{app:alg} includes a comprehensive description of the AL pipeline. 

We propose a set of acquisition functions to guide the selective annotation of concept labels during training. Let the unlabeled pool be defined as $\mathcal{U} \subset [|\mathcal{D}_{\text{pool}}|] \times [K]$, representing the Cartesian product of instance indices and concept indices. Each instance index $i \in [1, \ldots, |\mathcal{D}_{\text{pool}}|]$ corresponds to a tuple $(x_i, y_i, y_i')$, and each concept index $k \in [1, \ldots, K]$ corresponds to the $k$-th concept. Thus, each pair $(i, k) \in \mathcal{U}$ identifies which label $s_i^k$ to query. During each learning episode, we select a batch of $B$ pairs $(i, k)$ from $\mathcal{U}$ according to an acquisition function $\mathcal{A}_\bullet(i, k)$, annotate the selected concept labels, and retrain the model. We consider the following acquisition strategies:

\paragraph{Random Selection} As a baseline acquisition strategy, we uniformly sample concept-query tuples:
\begin{equation}
    \mathcal{A}_{\text{random}} = \text{UniformSample}(\mathcal{U}).
\end{equation}
\paragraph{Concept Variance} We select the concept-query tuples with the highest predictive variance in concept differences:
\begin{equation}
    \mathcal{A}_{\text{Var}}(i, k) = \text{Var}\left[\Delta c^k_i\right] = \sigma_k^2(x_i, y_i) + \sigma_k^2(x_i, y_i'),
\end{equation}
where $\Delta c^k_i$ is the $k$-th dimension of the concept difference.

\paragraph{Concept-weighted Influence Score (CwIS)}  
We select concept-query tuples based on their influence on the reward difference, weighted by how uncertain the model is about the prediction itself:
{\small
\begin{align}
    \mathcal{A}_{\text{CwIS}}(i, k) =\ & \left| r(x_i, y_i) - r(x_i, y_i') - r^{(k)}(x_i, y_i) + r^{(k)}(x_i, y_i') \right| \nonumber \\
    &\! + \lambda \cdot \text{Var}[\Delta c^k_i], 
\end{align}
}
where $r^{(k)}(\cdot)$ denotes the reward computed after intervening on the $k$-th concept (i.e., setting its logit to a fixed high/low value), and we set $\lambda$ to 0.1. This acquisition function targets concept annotations that are both highly influential for reward prediction and uncertain. Our CwIS strategy is inspired by the CooP policy of~\citet{chauhan2023interactive}, which combines concept uncertainty and influence for test-time interventions. While they target inference-time interaction, CwIS is the closest adaptation of existing complementary literature, despite differing goals.

\paragraph{Expected Information Gain (EIG)} EIG maximizes the expected reduction in uncertainty about the model's predictions after observing a concept label, which can be approximated by computing the difference between the expected entropy and the entropy of the expected prediction after sampling~\citep{houlsby2011bayesian}. Concretely, we have: 
\begin{align}
\mathcal{A}_{\text{EIG}}(i, k) &= \mathbb{E}_{\theta, \phi}\left[\mathcal{H}\left[p(s^k_i \mid x_i, y_i, y_i', \theta, \phi)\right]\right] \nonumber \\
&- \mathcal{H}\left[\mathbb{E}_{\theta, \phi}\left[p(s^k_i \mid x_i, y_i, y_i', \theta, \phi)\right]\right].
\end{align}

\section{Experimental Setup}

We evaluate our method on the UltraFeedback dataset~\citep{cui2024ultrafeedback}, a large-scale resource of diverse prompt-response pairs from 17 language models, designed to support alignment research without relying on explicit human preference labels. 
Further explanation of the dataset is included in Appendix~\ref{app:UF}.  To encode the prompts and responses, we leverage the representations predicted for an LLM encoder (See Figure~\ref{fig:overview}). In particular, to avoid data leakage, all embeddings are computed using LLaMA-2 7B~\citep{meta2023llama2}, whose weights were released prior to the UltraFeedback dataset. This ensures no overlap between model pretraining and evaluation data. To acquire the ground truth concept annotations resembling the human concept-preferences, we annotate each sample using an LLM judge, OpenAI GPT-4o-based, with ten interpretable and broadly applicable concepts: helpfulness, correctness, coherence, complexity, verbosity, instruction following, truthfulness, honesty, safety, and readability. These were chosen for their relevance to human evaluative reasoning. With our approach, we circumvent some of the potential issues present in prior work on interpretable reward models \citep{wang2024interpretable}. These include: repeated concept definitions, unevenly distributed annotations across datasets, potentially introducing bias \citep{kobalczyk2025preference}, the use of LLM encoders already trained on the same preference data, raising concerns of information leakage and confounding effects, the lack of concept-level evaluation, and finally, the assumed access to all concept labels. In contrast, our approach generates interpretable labels that are uniformly available across all data points, ensuring greater robustness for downstream learning. Further implementation details are found in Appendix~\ref{app:exp_det} and the \hyperlink{https://github.com/sonialagunac/cb-rm-workshop}{code}\footnote{\scriptsize \url{https://github.com/sonialagunac/cb-rm-workshop}} is publicly available. 

\vspace*{-0.15cm}
\section{Results}

\begin{figure}[h]
\centering
\setlength{\tabcolsep}{1pt} 
\renewcommand{\arraystretch}{1.0} 

\begin{tabular}{cc}
\includegraphics[width=0.5\linewidth]{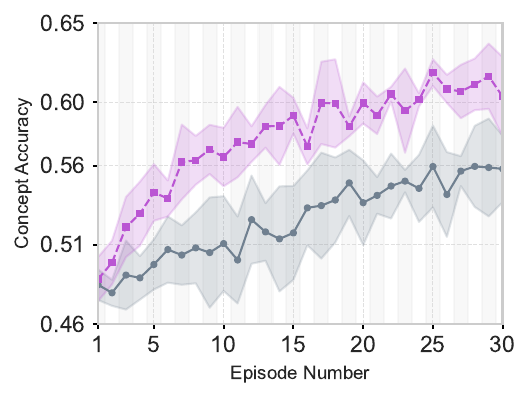} &
\includegraphics[width=0.5\linewidth]{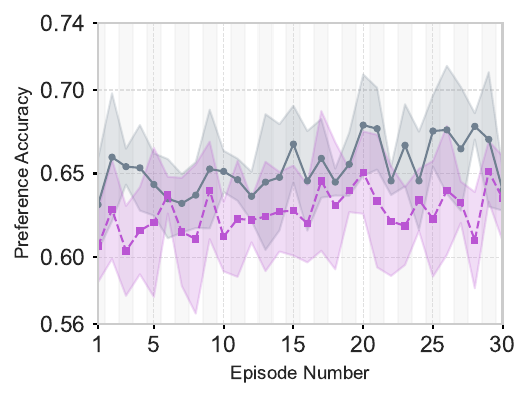} \\
\includegraphics[width=0.5\linewidth]{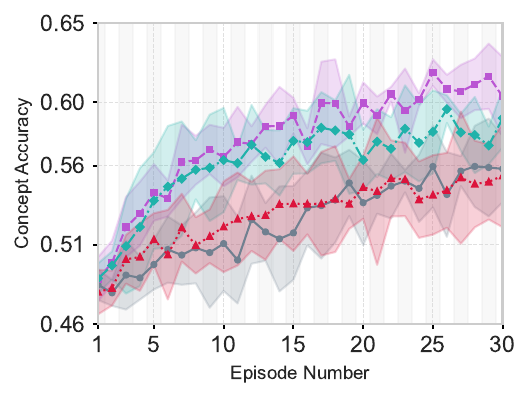} &
\includegraphics[width=0.5\linewidth]{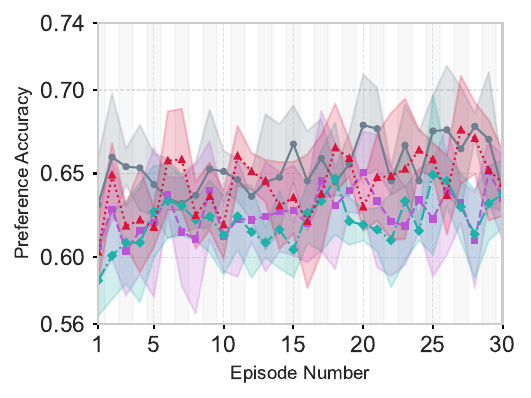} \\
\multicolumn{2}{c}{\includegraphics[width=\linewidth]{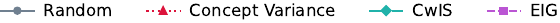}} \\
\end{tabular}
\vspace*{-0.5cm}
\caption{
\textbf{Top:} Concept and preference accuracy in the EIG and random acquisition strategies. 
\textbf{Bottom:} Concept and preference accuracy in all studied acquisition strategies.
Shaded areas denote standard deviation across 5 seeds.
}
\label{fig:main_results}
\vspace{-1em}
\end{figure}

To assess the effectiveness of the proposed acquisition functions, we track the improvement in concept accuracy across AL episodes, aiming to enhance reward model interpretability in a cost-efficient manner. In Figure~\ref{fig:main_results} (Top), we see how EIG consistently achieves the fastest gains in concept accuracy when compared with the random baseline. In Figure~\ref{fig:main_results} (Bottom), we include a comparison with the remaining baselines (not in (Top) for clarity), showing that CwIS performs closer to EIG, while concept variance does not significantly affect. Each plot is evaluated over five random seeds. At the same time, preference accuracy remains comparable across all acquisition methods.

\section{Discussion and Conclusion}
Although preference accuracy remains comparable across methods, models trained with EIG are notably more robust and interpretable in terms of concept performance  (Figure~\ref{fig:main_results}). Among baselines, CwIS ranks second after EIG. However, unlike in test-time intervention methods~\citep{chauhan2023interactive,shin2023closer}, concept uncertainty alone does not ensure optimal active learning. This underscores a key distinction: intervention targets immediate fixes, while active learning must prioritize long-term representation quality and generalization. The instability in preference performance likely stems from noisy or biased feedback~\citep{casper2023open,sharma2024towards}, especially for complex responses, leading to misaligned or saturated reward signals.

We also observe that embeddings from LLMs that have been likely exposed to the UltraFeedback dataset during their pretraining, exhibit strong concept correlations at the representation level, and the additional concept-level supervision does not improve performance (more details in Appendix~\ref{more_results}). This observation suggests a potential information leakage in the LLM's representations. Our findings caution against blind reliance on encoder outputs and suggest a direction for future work on identifying the impact of data leakage in robust reward modeling.

\vspace*{-0.2cm}
\paragraph{Conclusions} We propose CB-RM, a concept-based reward modeling framework that enables interpretable preference learning through selective concept annotation. By introducing an active learning formulation and an EIG-based acquisition strategy, we show that it is possible to efficiently improve concept accuracy in low-supervision settings without sacrificing preference performance. 
Our results highlight EIG’s effectiveness in aligning latent concepts with human preferences, paving the way for more transparent and auditable reward models. While challenges and open questions remain, our approach—combining CBMs, information-theoretic acquisition, and structured training—presents a step toward \textit{programmatic reward models} aligned with human reasoning.

\section*{Acknowledgments}

This work was supported by Azure sponsorship credits granted by Microsoft’s AI for Good Research Lab. SL is supported by the Swiss State Secretariat for Education, Research, and Innovation (SERI) under contract number MB22.00047. KK is supported by funding
from Eedi.

\bibliographystyle{apacite}
\bibliography{bibliography}

\newpage

\onecolumn
\appendix

\section{Active Learning in CB-RM}
\label{app:alg}
\vspace*{-0.15cm}
Algorithm~\ref{alg:active_cbm} outlines the active learning loop used in CB-RM, including acquisition, labeling, and model retraining steps.

\begin{algorithm}[h!]
\caption{Active Learning with CB-RM}
\label{alg:active_cbm}
\small
\begin{algorithmic}[1]
\REQUIRE Pool dataset $\mathcal{U}$, labeled set $\mathcal{L}$, number of episodes $T$, number of labels to query at each episode $B$.
\STATE Train model on initial labeled set $\mathcal{L}$
\FOR{episode $t = 1$ to $T$}
    \STATE Compute acquisition scores for all candidate $(x,y) \in \mathcal{U}$ using an acquisition function
    \STATE Select $B$ indices $(i, k)$ to query the concept annotations.
    \STATE For each pair $(i,k)$, query the $k$-th concept label of the $i$-th instance $(x_i, y_i, y_i')$ and update the labeled set $\mathcal{L}$.
    \STATE Remove selected indices from the pool $\mathcal{U}$.
    \STATE Retrain CB-RM model on the updated labeled set $\mathcal{L}$.
\ENDFOR
\end{algorithmic}
\end{algorithm}

\vspace*{-0.3cm}
\section{Experimental Details}
\subsection{Ultrafeedback Dataset}
\label{app:UF}
We use the UltraFeedback dataset~\citep{cui2024ultrafeedback}, a large-scale resource of 63,967 prompts and 255,864 responses designed for training and evaluating alignment methods in large language models. Prompts are sourced from high-quality datasets like TruthfulQA, UltraChat, FLAN, and ShareGPT, and span a broad range of tasks including question answering, instruction following, and factual verification. Each prompt is paired with four responses sampled from a diverse pool of 17 open-source and commercial models (i.e., LLaMA2, GPT-4, Vicuna), ensuring broad stylistic and qualitative coverage.

Responses were generated using varied decoding strategies and model sizes to capture natural variation in assistant behavior. The instructions are primarily single-turn queries but include both simple factual and more complex creative or ethical tasks. This diversity makes UltraFeedback a robust testbed for preference modeling without relying on explicit human labels, offering rich signal for evaluating interpretable, concept-based reward models.
\vspace*{-0.2cm}
\subsection{Extended Implementation Details}
\label{app:exp_det}
Regarding the dataset, we use a train-validation-test split of 70-10-20 of Ultrafeedback. In order to generate the concept annotations using LLM judge, for each example, the LLM was presented with a system prompt and a user query, followed by two alternative assistant responses. The model was then instructed to rate which response was better with respect to ten predefined concepts. Each concept was scored independently on a scale from 0 to 1, where a score of 0 indicates that the first response is clearly better, and 1 that the second response is clearly better. We generate the final preference labels as a linear combination of the mentioned concepts to ensure the ultimate reward model remains interpretable. 

Moreover, the implemented FIFO has a buffer capacity of 32000 and a number of acquired samples of 320. Both the CBM and gating mechanism use a one-layer perceptron to extract concept logits in a probabilistic variant and weights, and were trained for 1 epoch. This architectural modeling ensures the use of programmatic representations: by structuring reward models through interpretable components, we enable modular, debuggable learning that goes beyond black-box supervision. 
\vspace*{-0.5cm}
\section{Active Learning Results on Large LLMs with Information Leakage}
\label{more_results}
\vspace*{-0.1cm}
\begin{figure}[h!]
\centering
\setlength{\tabcolsep}{1pt} 
\renewcommand{\arraystretch}{1.0} 
\vspace*{-0.2cm}
\begin{tabular}{cc}
\includegraphics[width=0.3\linewidth]{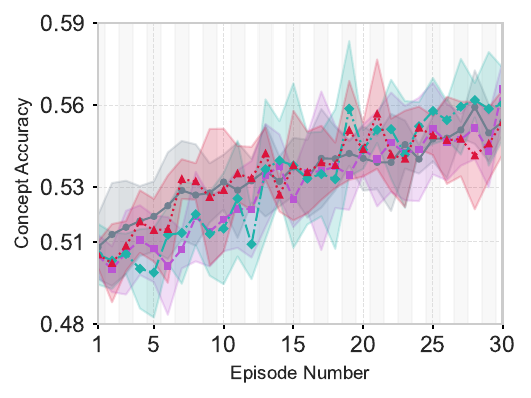} &
\includegraphics[width=0.3\linewidth]{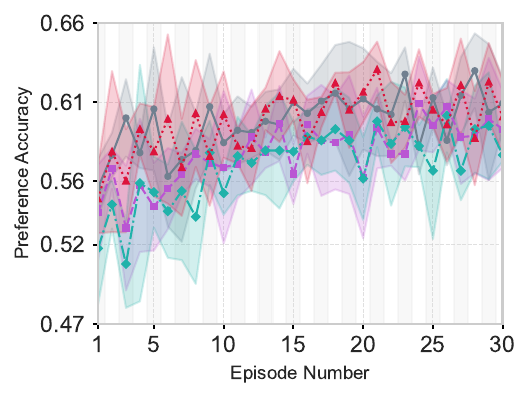} \\
\multicolumn{2}{c}{\includegraphics[width=0.48\linewidth]{figures/legend.pdf}} \\
\end{tabular}
\caption{
Concept and preference accuracy in all studied acquisition strategies using embeddings generated from Llama3-8B.
Shaded areas denote standard deviation across 5 seeds.
}
\label{fig:main_results2}
\end{figure}
Figure~\ref{fig:main_results2} illustrates the limitations of applying active learning with concept-based acquisition strategies on large LLMs such as LLaMA-3-8B, whose training cutoff date is after the release of the UltraFeedback dataset and likely have been exposed to this dataset during their pre-training. In this setting, embeddings exhibit strong linear correlations with target concepts, and additional supervision yields little to no improvement. This highlights a critical risk of information leakage, where pretrained representations already encode the target signals, undermining the benefits of active acquisition. We include this analysis to caution practitioners: when evaluating interpretability or data efficiency methods, care must be taken to avoid confounded setups that mask true learning dynamics. Furthermore, the CBM paradigm itself has been shown to be vulnerable to information leakage within the bottleneck, with recent works proposing methods to quantify this effect~\citep{makonnen2025measuring}. Therefore, it is crucial to avoid additional sources of leakage in encoder representations, which would only compound the problem and obscure the model's reasoning further.

\end{document}